\documentclass{article}

\usepackage{PRIMEarxiv}
\usepackage{xcolor}
\usepackage{verbatim}
\usepackage{amsmath} 
\usepackage{bm}        % ← ADD THIS LINE
\usepackage[utf8]{inputenc} % allow utf-8 input
\usepackage[T1]{fontenc}    % use 8-bit T1 fonts
\usepackage{hyperref}       % hyperlinks
\usepackage{booktabs}       % professional-quality tables
\usepackage{amsfonts}       % blackboard math symbols
\usepackage{nicefrac}       % compact symbols for 1/2, etc.
\usepackage{microtype}      % microtypography
\usepackage{lipsum}
\usepackage{subcaption} % for subfigure environment
\usepackage{fancyhdr}       % header
\usepackage{graphicx}       % graphics
\graphicspath{{media/}}     % organize your images and other figures under media/ folder

\title{Diluting Restricted Boltzmann Machines}

\author{
  C. Díaz-Faloh and R. Mulet \\
  Group of Complex Systems and Statistical Physics, Department of Theoretical Physics,\\
  Physics Faculty, University of Havana, Cuba\\
  \texttt{cristina.diaz@fisica.uh.cu} 
}

\begin{document}
\maketitle

\begin{abstract}
Recent advances in artificial intelligence have relied heavily on increasingly large neural networks, raising concerns about their computational and environmental costs. This paper investigates whether simpler, sparser networks can maintain strong performance by studying Restricted Boltzmann Machines (RBMs) under extreme pruning conditions. Inspired by the Lottery Ticket Hypothesis, we demonstrate that RBMs can achieve high-quality generative performance even when up to 80\% of the connections are pruned before training, confirming that they contain viable sub-networks. However, our experiments reveal crucial limitations: trained networks cannot fully recover lost performance through retraining once additional pruning is applied. We identify a sharp transition above which the generative quality degrades abruptly when pruning disrupts a minimal core of essential connections. Moreover, re-trained networks remain constrained by the parameters originally learned  performing worse than networks trained from scratch at equivalent sparsity levels. These results suggest that for sparse networks to work effectively, pruning should be implemented early in training rather than attempted afterwards. Our findings provide practical insights for the development of efficient neural architectures and highlight the persistent influence of initial conditions on network capabilities.
\end{abstract}

% keywords can be removed

\section{Introduction}
Over the past decade, Artificial Intelligence (AI) and Machine Learning (ML) have developed rapidly, yielding impressive results in complex tasks such as object recognition \cite{krizhevskyImageNetClassificationDeep2017}, language synthesis \cite{ParallelWaveNetFast}, and protein structure prediction \cite{jumperHighlyAccurateProtein2021}. This excitement has extended beyond the research community to the general public through tools like ChatGPT \cite{openaiGPT4TechnicalReport2023} and more recently DeepSeek \cite{deepseek}.

This progress is closely linked to the increasing availability of large datasets and computing power. The widespread use of the internet and the automation of many human activities have made vast amounts of data accessible, while advances in hardware have allowed for the training of increasingly large and complex models. As a result, the field has largely focused on scaling: using deeper networks, larger datasets, and more intensive training algorithms to improve performance. However, it is unlikely that this approach can continue indefinitely. At some point, technical limitations will arise and the environmental impact of large-scale training, already significant, will become of even greater concern \cite{dharCarbonImpactArtificial2020,strubellEnergyPolicyConsiderations2019}. In response, the community has shown renewed interest in understanding what neural networks learn, how they do it, and which elements are truly essential.

In this context, the Lottery Ticket Hypothesis, introduced by Frankle and Carbin in 2019 \cite{lotTicket1}, offers a compelling perspective. They showed that a large, randomly initialized neural network often contains smaller sub-networks, called "winning tickets", that, when trained in isolation with their original initialization, can match or exceed the performance of the full model. This surprising result suggests that much of the model’s capacity may be superfluous, and that identifying these sub-networks could lead to more efficient learning. Follow-up work has generalized and refined these ideas \cite{lotTicket2,lotTicket3}. 

This work draws inspiration from the Lottery Ticket Hypothesis and seeks to investigate the robustness of a simple generative neural network, the Restricted Boltzmann Machine (RBM), under extreme pruning conditions. Specifically, our goal is to examine how well pruned RBMs can retain generative performance, and whether suitable retraining strategies can restore or even improve it. We train RBMs, prune a significant portion of their connections, apply different retraining protocols, and evaluate the quality of their generative output. Our results aim to shed light on the relationship between sparsity, initialization, and generalization in these networks.

The paper is organized as follows: we begin with a brief theoretical overview RBMs, followed by a section that defines the architecture and metrics employed in our experiments. Next, we present the experimental setup and results, concluding with a discussion of the findings.

\section{Brief Overview of RBMs}\label{sec:intro}

A Restricted Boltzmann Machine (RBM) \cite{item_1,item_2,item_3,asja} is a Markov random field with interactions defined on an undirected bipartite graph. The graph consists of a layer of $m$ visible neurons $\bm{V}=(V_1,...,V_m)$ and a second layer of $n$ hidden neurons $\bm{H}=(H_1,...,H_n)$. The visible neurons correspond to the observable data, while the hidden neurons capture dependencies between the observed variables. RBMs are "restricted" in comparison with Boltzmann Machines because there are no intra-layer connections. The energy function for a configuration $(\bm{v},\bm{h})$ is defined as:

\begin{equation}
  E(\bm{v},\bm{h})=-\sum_{i,j}h_iW_{ij}v_j-\sum_{i}h_ic_i-\sum_jv_jb_j
  \label{eq:hamiltonian}
\end{equation}

where $w_{ij}$ is the weight between visible unit $v_j$ and hidden unit $h_i$, and $b_j$, $c_i$ are bias terms. The joint probability distribution follows the Boltzmann distribution:

\begin{equation}
p(\bm{V},\bm{H})=\dfrac{e^{-E(\bm{V},\bm{H})}}{Z}, \quad Z=\sum_{\bm{V},\bm{H}}e^{-E(\bm{V},\bm{H})}
\end{equation}

Training an RBM involves adjusting the parameters $\bm{\theta} = (\bm{W}, \bm{b}, \bm{c})$ to maximize the likelihood of the training data. The gradient of the log-likelihood leads to expressions where each term contains the difference between data-dependent and model-dependent expectations:

\begin{equation}
\frac{\partial \ln \mathcal{L}(\bm{\theta}|\bm{v})}{\partial b_j}
= 
\langle v_j \rangle_{\text{data}}
- 
\langle v_j \rangle_{\text{model}},
\end{equation}
\begin{equation}
\frac{\partial \ln \mathcal{L}(\bm{\theta}|\bm{v})}{\partial W_{ij}}
=
\langle h_i v_j  \rangle_{\text{data}}
-
\langle h_i v_j  \rangle_{\text{model}},
\end{equation}
\begin{equation}
\frac{\partial \ln \mathcal{L}(\bm{\theta}|\bm{v})}{\partial c_i}
=
\langle h_i \rangle_{\text{data}}
-
\langle h_i \rangle_{\text{model}}.
\end{equation}

The model expectations (negative phase terms) in the RBM gradient require sampling from the model distribution, which is computationally intractable for most practical network sizes. In practice, approximate methods are used to estimate these expectations. Contrastive Divergence (CD) \cite{item_4} and Persistent Contrastive Divergence (PCD) \cite{item_5} are two commonly used techniques based on Gibbs sampling. CD initializes chains from training data and performs a few steps of alternating sampling, while PCD maintains persistent chains across parameter updates. Both methods rely on the assumption that even short runs of Gibbs sampling can provide useful information for learning, offering a reasonable trade-off between accuracy and efficiency.

%Because of their relatively simple structure and generative capabilities, RBMs provide a well-suited model for investigating how pruning affects learning and generation. In the following section, we describe the architecture and training setup used in our experiments.

\section{Model and Evaluation Framework}\label{sec:model-eval} %-----------------------------------------------------------------------------------

To investigate the impact of pruning on learning and generation, we begin by training a baseline Restricted Boltzmann Machine (RBM) on a reference dataset. The goal is to ensure that we start from a consistently trained model so that any variations in performance can be reliably attributed to the pruning procedure rather than to training instability or randomness.

We use a standard single-layer RBM and train it on MNIST \cite{mnist}, a widely used dataset composed of 60,000 gray-scale images of handwritten digits with a balanced representation across the ten digit classes. The architecture of our model reflects the input image dimensions, with 784 visible units (28$\times$28 pixels) and 400 hidden units, resulting in 313,600 trainable weights. Training is performed using Persistent Contrastive Divergence (PCD) with 1,000 Gibbs sampling steps per iteration and run training for 10,000 iterations, monitoring the pseudo-likelihood to assess convergence.

\begin{figure*}[!hb]%
\centering
\begin{subfigure}{.3\textwidth}
  \includegraphics[keepaspectratio=true,width=0.9\textwidth]{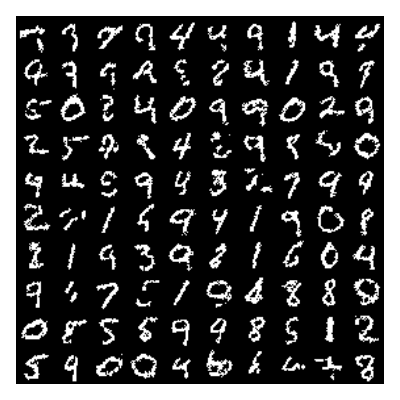}
  \caption{}
\end{subfigure}%
\begin{subfigure}{.3\textwidth}
  \centering
  \includegraphics[keepaspectratio=true,width=0.9\textwidth]{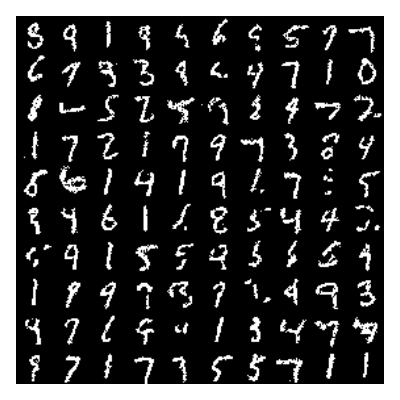}
  \caption{}
\end{subfigure}
\begin{subfigure}{.3\textwidth}
  \centering
  \includegraphics[keepaspectratio=true,width=0.9\textwidth]{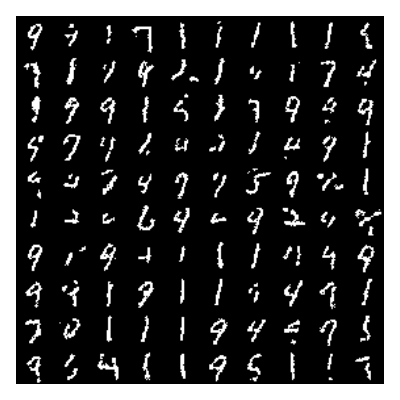}
  \caption{}
\end{subfigure}
\caption{(a) Examples generated by an RBM trained without any initial dilution. (b) Additional pruning of $p=20\%$ applied. (c) Additional pruning of $p=30\%$ applied. It can be observed that the generated images lose diversity as the additional dilution increases.}%
\label{gend}
\end{figure*}

To ensure statistical robustness, we train multiple independent replicas of the model (15 in each case). Each replica is initialized independently and trained with a different random order of the dataset. This allows us to distinguish systematic effects of pruning from normal fluctuations in the learning trajectory. The trained models demonstrate competent generative behavior, producing plausible samples of handwritten digits (see Figure~\ref{gend}).

Once trained, we proceed to evaluate the generative quality of the models under various pruning conditions. However, quantifying generation quality is not a trivial task as the true distribution underlying the training data is often-and certainly in this case-unknown. Moreover, there are multiple aspects to consider when defining what makes a good generated sample. For instance, in a dataset composed of diverse digits, one would not only want the generated samples to resemble plausible digits but also to reflect the diversity present in the training set.

Due to this complexity, we employ three different metrics to assess generation quality and to verify whether the qualitative results remain consistent across them. These are the Second Moment Error (E$^{(2)}$) and the Adversarial Accuracy Indicator Error (E$_{AAI}$), previously used in \cite{aurelien} and \cite{EAA}, respectively, along with a more heuristic method introduced by us in this work, which aims to emulate human evaluation by means of an additional model. A brief description of these metrics follows:

\textbf{Second Moment Error (E$^{(2)}$)}: 
This measures the mean squared error (MSE) between the covariance matrices computed from RBM-generated samples and the training dataset. It quantifies how well the model captures pairwise feature correlations in the data.

\textbf{Adversarial Accuracy Indicator Error (E$_{AAI}$)}: 
This metric evaluates both the similarity and uniqueness of the data generated relative to the training set. The computation involves:
\begin{itemize}
\item Two sample sets: target $T = \{\mathbf{v}_{RBM}^{(m)}\}_{m=1}^{N_s}$ (generated) and source $S = \{\mathbf{v}_{D}^{(m)}\}_{m=1}^{N_s}$ (dataset)
\item Four nearest-neighbor distances for each sample $m$:
\begin{itemize}
\item $d_{TS}(m) = \min_n \|\mathbf{v}_{RBM}^{(m)} - \mathbf{v}_{D}^{(n)}\|$
\item $d_{ST}(m) = \min_n \|\mathbf{v}_{D}^{(m)} - \mathbf{v}_{RBM}^{(n)}\|$
\item $d_{SS}(m)$, $d_{TT}(m)$ (within-set distances)
\end{itemize}
\item The E$_{AAI}$ is the MSE between the observed nearest-neighbor frequencies and the ideal 0.5 value expected for indistinguishable distributions
\end{itemize}

\textbf{Additional Classifier}: We trained an additional model to classify the generated samples as either digits or non-digits. The confidence it assigns to a sample being a digit is taken as a measure of generation quality. This model also includes an extra layer that, once a sample is classified as a digit, determines to which of the ten digit categories it belongs. Given that MNIST has a balanced distribution across the ten digits, one can assess how far the generated samples deviate from the ideal uniform distribution using
\begin{equation}
    d_f=\frac{1}{10}\sum_{i=0}^9\left(f_i-\frac{1}{10}\right)^2
\end{equation}
where $i$ indexes the categories of digits and $f_i$ is the frequency with which the digit $i$ is generated.

In the end, the additional classifier provides two useful measures: a ``generation quality'' score, which is higher when individual samples resemble well-written digits, and the value $d_f$, which captures the diversity of the generated digits and equals zero in the optimal case. This method has the advantage of mimicking how a human would intuitively assess the quality of the generated digits, making it more interpretable. Moreover, training the additional model is straightforward and computationally inexpensive. More details on the architecture of this auxiliary network are provided in Appendix~(\ref{appendix:negative samples}).

%Although the method offers interpretability it also introduces an additional black box into the system. Furthermore, it is highly tailored to the specific conditions of our experiments, limiting its generalizability to other scenarios. For this reason, it is important to combine multiple evaluation metrics to draw robust conclusions from the experiments.
  
\section{Experiments}
\label{sec:experiments}

\subsection{Adaptation to Initial Pruning and Breakdown under Further Sparsification}
\label{subsection:exp1}
In our first experiment, we explore how restricting the connectivity of the network from the outset affects learning and generative performance. To do this, we randomly select a fraction $p_0$ of the weights $w_{ij}$ and set them to zero immediately after initialization, effectively removing those connections before any training begins. Throughout the entire training process, we keep this subset of weights fixed at zero, ensuring that they remain disconnected. It is important to emphasize that this random disconnection of individual weights is not equivalent to removing entire hidden units from the network; the latter would impose a more severe constraint. Instead, our procedure imposes sparsity while preserving the distributed representation capabilities of the RBM. Once the network has been trained under this initial sparsity constraint, we generate samples and evaluate their quality using the metrics described in the previous section. 

\begin{figure}[!ht]%
    \centering
    \includegraphics[width=0.5\linewidth]{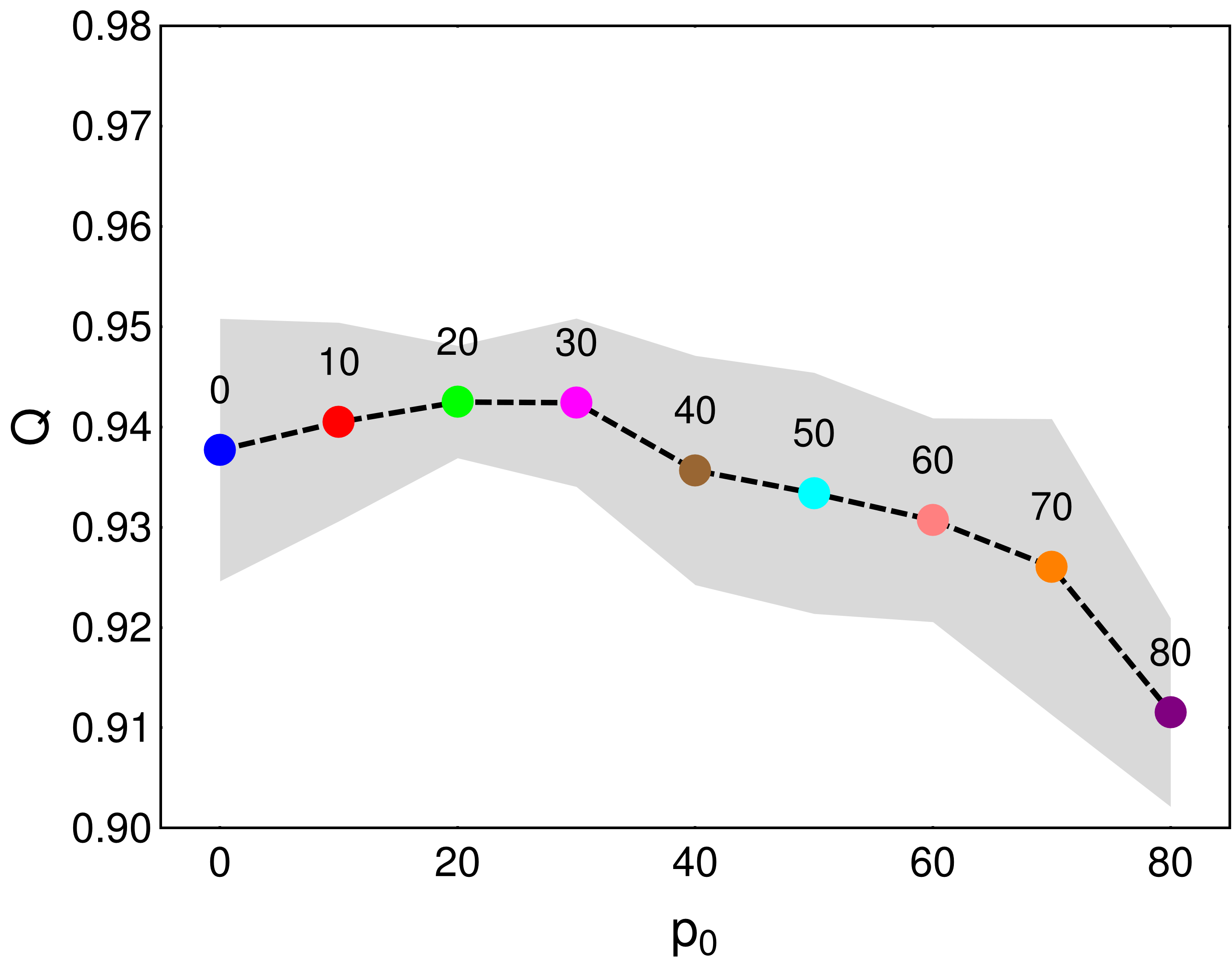}
    \caption{Generation quality $Q$ assigned by the additional classifier as a function of the initial pruning fraction $p_0$ used during training. Note that the $Q$ axis starts at $0.90$, so the apparent downward trend is less pronounced than it might initially seem.}
    \label{fig:q0}
\end{figure}

Figure~(\ref{fig:q0}) shows the quality of the generation $Q$, as measured by the auxiliary classifier, for different values of the initial dilution $p_0$. There is a slight downward trend in quality as $p_0$ increases, but the effect is small. All the models, including those trained with up to 80\% of their weights removed, consistently produce high-quality samples. This indicates that the initial sparsity does not compromise the model’s ability to learn the generative structure of the data.

This observation supports the well-known idea that modern neural networks often operate in an over-parameterized regime, where only a subset of weights is actually needed to achieve good performance. It also resonates with the Lottery Ticket Hypothesis, which proposes that successful training depends on the presence of a small sub-network—“the winning ticket”—within a larger model. Although setting $80\%$ of the weights to zero before training might seem extreme, the large number of parameters in our RBMs means that the probability of initializing such a winning ticket remains high. As long as one of these sub-networks survives the initial pruning, the model can still learn effectively and produce good samples.

To investigate the resilience of the trained network to additional sparsification, we perform a second stage of pruning. This additional sparsification is designed to test how the generative performance degrades as more and more information is removed from the weights. Starting from already trained RBMs, with different initial levels of dilution $p_0$, we progressively turn off the remaining smallest non-zero weights, and after each pruning step, we evaluate the quality of the generated samples. This process allows us to study how robust the network remains as connectivity is reduced. It is important to emphasize that this second pruning stage differs fundamentally from the initial dilution $p_0$: while $p_0$ defines the sparsity pattern before and throughout training, the additional pruning $p$ is purely a post-training intervention. It consists solely of setting a fraction of the remaining weights to zero, without any further optimization, gradient calculations, or retraining.

\begin{figure*}[ht!]%
\centering
\begin{subfigure}{.45\textwidth}
  \includegraphics[keepaspectratio=true,width=0.9\textwidth]{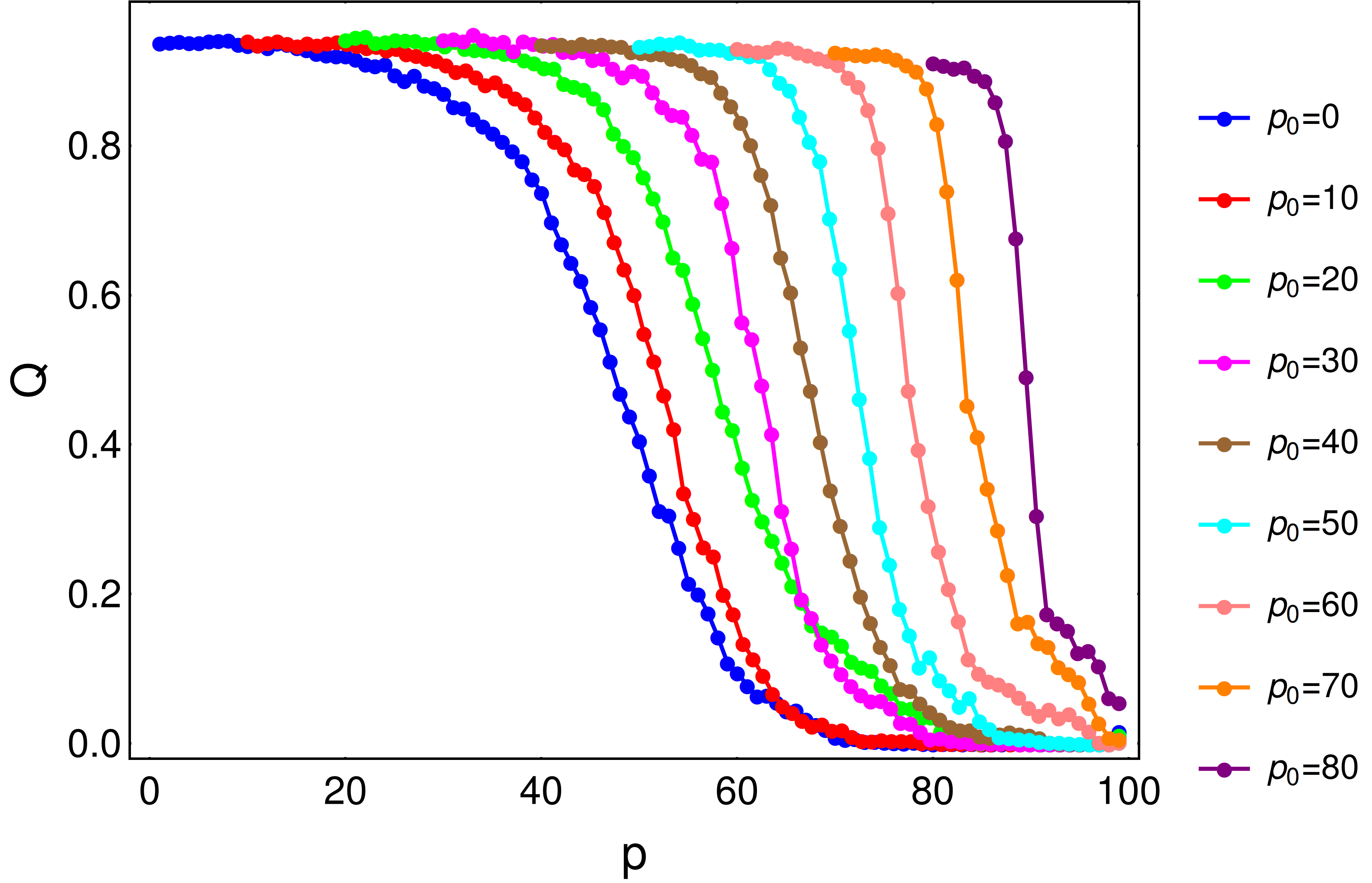}
  \caption{}
  \label{quality}
\end{subfigure}%
\begin{subfigure}{.45\textwidth}
  \centering
  \includegraphics[keepaspectratio=true,width=0.9\textwidth]{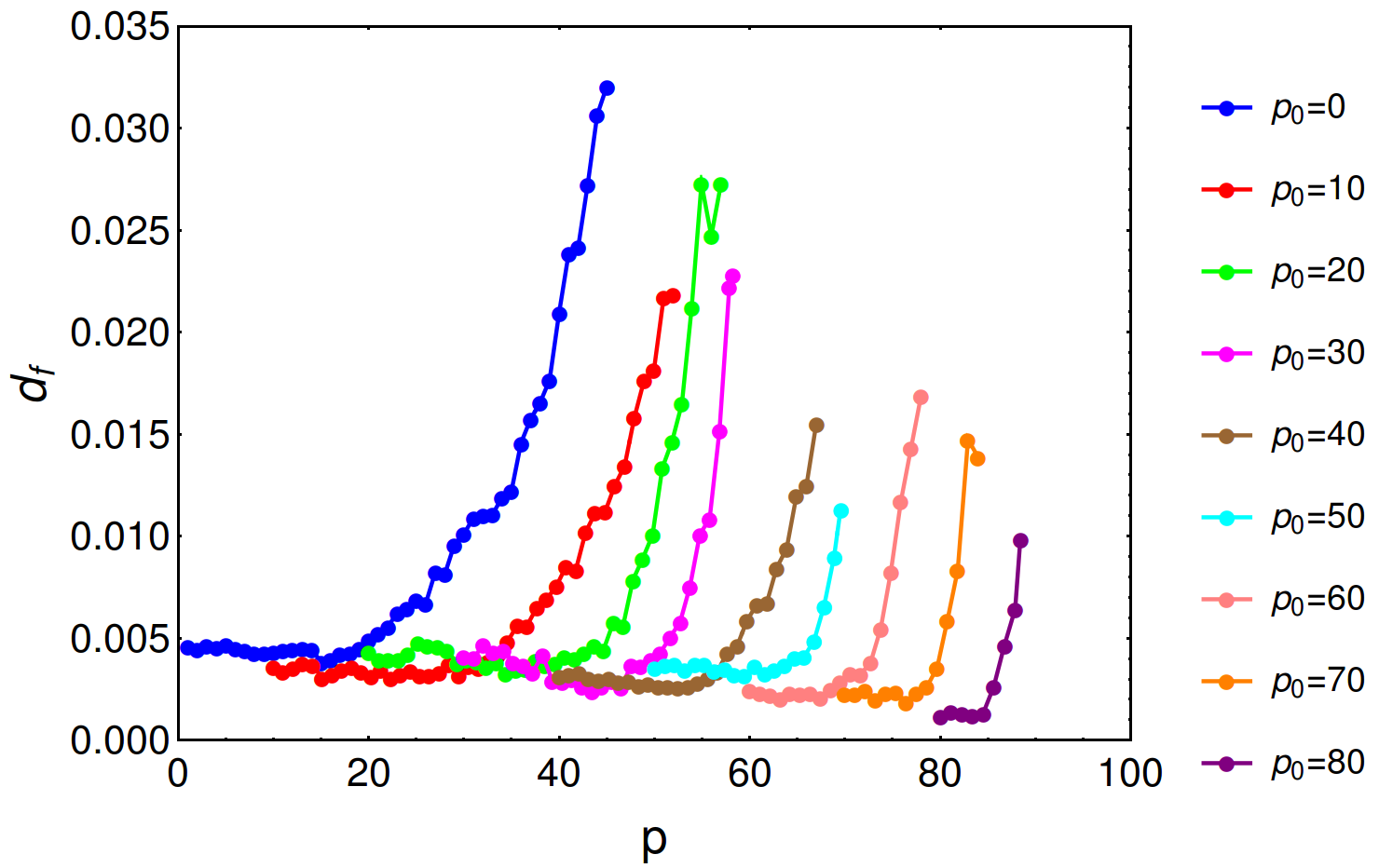}
  \caption{}
  \label{diversidad}
\end{subfigure}
\begin{subfigure}{.45\textwidth}
  \centering
  \includegraphics[keepaspectratio=true,width=0.9\textwidth]{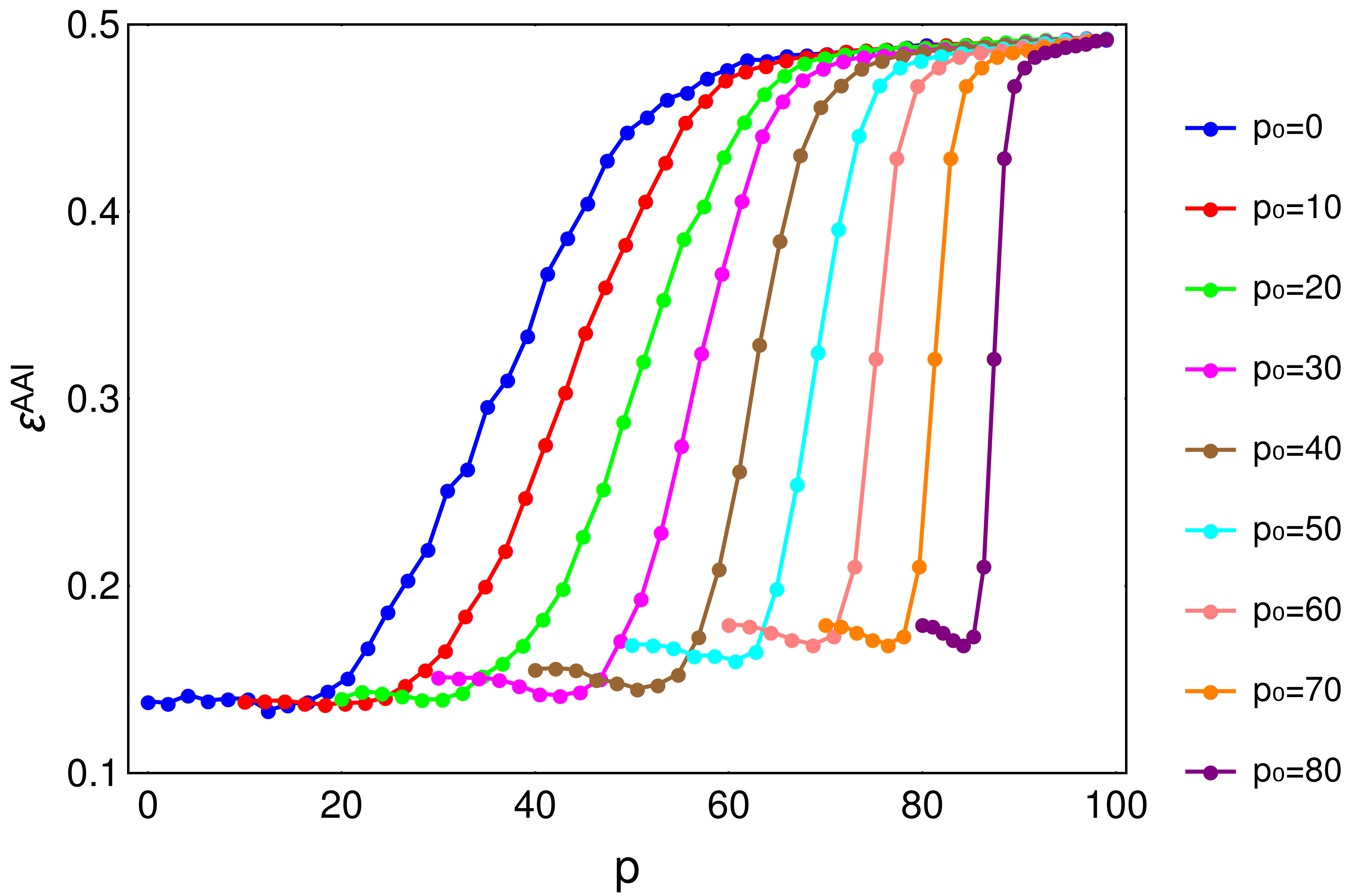}
  \caption{}
  \label{EAA}
\end{subfigure}
\begin{subfigure}{.45\textwidth}
  \centering
  \includegraphics[keepaspectratio=true,width=0.9\textwidth]{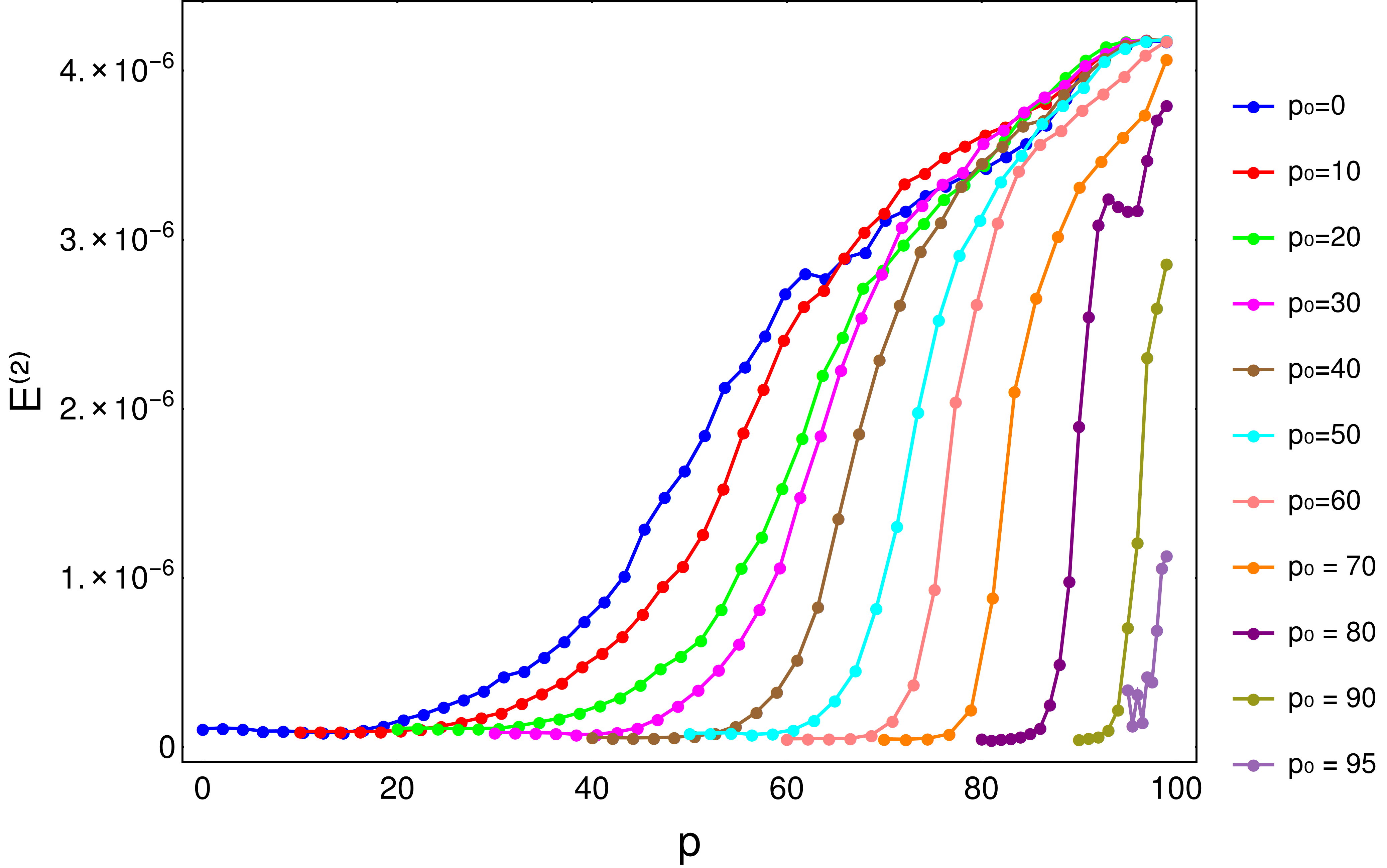}
  \caption{}
  \label{E22}
\end{subfigure}
\caption{(a)Generation quality of the RBMs given by the additional model (b) The curves show the distance $d_f$ between the uniform distribution of the 10 MNIST digits and the frequency of generation of each digit in the samples generated by RBMs with initial dilution $p_0$. On the x-axis, the additional pruning applied to the replicas is represented, and the curves are truncated at the value of $p$ where the generation quality drops to zero. Since it is meaningless to distinguish digits in images that are no longer classifiable as such, truncation is necessary. (c)Adversarial error EAA. (d)Error of the second moment.}%
\end{figure*}

The results of this experiment are presented in Figure~(\ref{quality}), which shows the quality of the generation as a function of the total pruning. The figure also displays the four metrics described previously. Since all behaviors are consistent, unless something else is mentioned, we will focus our disussion on $Q$. Each curve corresponds to a model trained with a different dilution $p_0$, ranging from $0\%$ to $80\%$ of the weights removed at initialization. As shown in Figure~(\ref{quality}), the generative quality remains high and nearly constant for small pruning $p$. However, when $p$ becomes large enough, the quality begins to drop rapidly and eventually stabilizes at very low values. This behavior reveals the existence of two distinct regimes: one in which the network generates good samples, and another in which its generative capacity breaks down abruptly. The sharpness of the transition suggests that there exists a relatively small set of crucial weights that contain the essential information required to produce meaningful samples. As long as most of these important connections are preserved, the network continues to function well; but then its performance decays abrouptly after a certain threshold is traspased. %once enough of them are removed, the generative ability collapses.

The point at which this breakdown occurs is not fixed across all models; it depends on the initial dilution $p_0$. Networks trained with higher $p_0$ values tend to withstand additional more aggressive pruning before their quality decreases. This indicates that the network learns to adjust to the constraints imposed during training, concentrating relevant information into the remaining weights. As a result, it is possible to obtain models that perform well even when a very large fraction of the weights (up to $80\%$) are turned off.

\begin{figure*}[ht!]%
\centering
\begin{subfigure}{.5\textwidth}
  \includegraphics[keepaspectratio=true,width=0.97\textwidth]{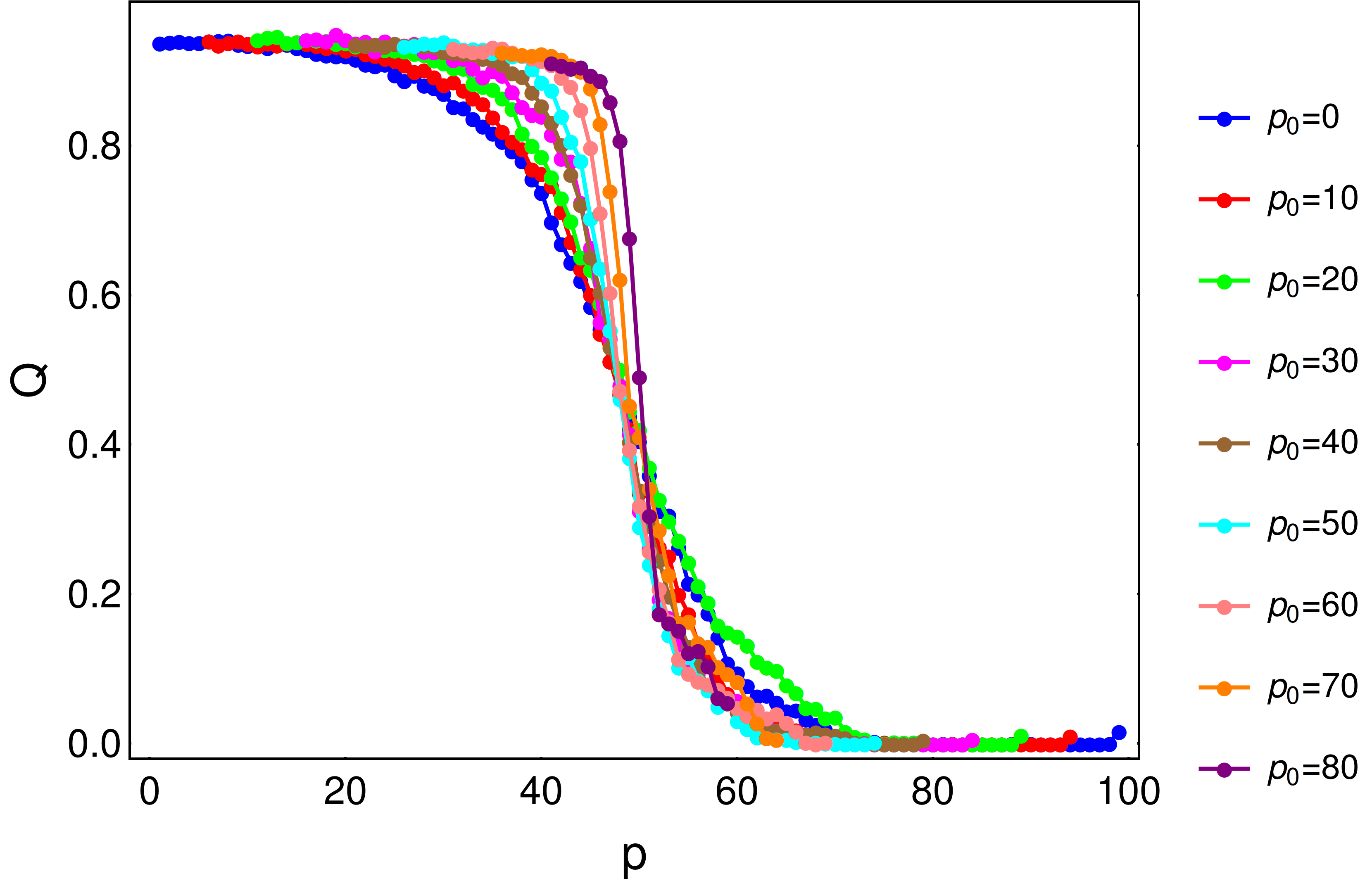}
  \caption{}
  \label{rescale}
\end{subfigure}%
\begin{subfigure}{.5\textwidth}
  \centering
  \includegraphics[keepaspectratio=true,width=0.83\textwidth]{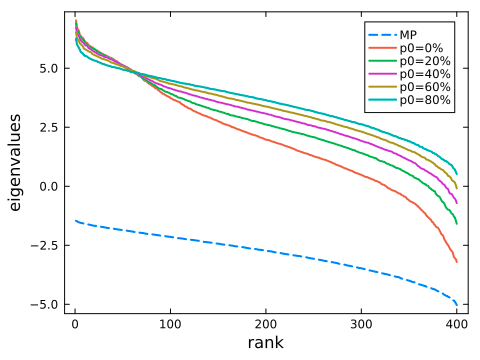}
  \caption{}
  \label{eigen}
\end{subfigure}
  \caption{(a)Rescaling of the generation quality curves for different $p_0$. The new independent variable is $p*=p-\alpha p_0$ with $\alpha=0.5$, which was the value that gave the best overlap between the curves. (b) The eigenvalues of networks with different dilution degrees are plotted in decreasing order, such that the largest eigenvalue has abscissa 1, the next 2, and so on. They are also
compared with the Marchenko-Pastur (MP) distribution. All the curves corresponding to the networks are cut off at the same point around
eigenvalue number 62.} 
\end{figure*}

This adaptive behavior is reflected by collapsing the curves. As seen in Figure~(\ref{rescale}), the curves corresponding to different $p_0$ values can be approximately follow a single universal curve if plotted against a rescaled pruning level $p^* = p - \alpha p_0$, with $\alpha = 0.5$. In this representation, the curves approach the behavior

\begin{equation}
 Q(p-\alpha p_0)= \begin{cases} 1, & \text{if}\ p-\alpha p_0<50 \\ 0, & \text{if}\ p-\alpha p_0>50 \end{cases} 
\end{equation},
 
suggesting that the quality of the generation depends primarily on the total effective pruning level, rather than on the individual contributions of $p_0$ and $p$. %This collapse highlights a simple, underlying principle governing the network’s capacity to tolerate sparsity.

To gain insight into how the network adapts to pruning, we analyze the spectrum of the weight matrices $w_{ij}$ after training. In Figure~(\ref{eigen}), we plot the eigenvalues in descending order for models trained with different initial sparsity levels $p_0$, and compare them to the Marchenko-Pastur distribution expected for random matrices. In all cases, the trained networks show significant deviations from the random expectation, even for high values of $p_0$. This indicates that training consistently introduces structure into the weights, regardless of the initial sparsity. Interestingly, all curves intersect around eigenvalue number 62, suggesting that the top 62 modes are preserved across different pruning conditions. These eigenvalues likely represent the most informative directions in weight space and may constitute the minimal spectral support required for successful data generation. On the other hand, the curves of the prunned RBM's, at larger eigenvalues, have the shape of the the Marchenko-Pastur (MP) distribution (see the blue dashed curve in  Figure~(\ref{eigen})). %in highly pruned networks reflects the gradual reduction of the learned structure as the number of parameters is aggressively reduced. {\bf I DON'T GET THE MEANING OF THIS SENTENCE. THE CURVES DON'T LOOK ALIKE}

\subsection{Retraining}

In the previous section, we saw that very diluted networks can generate images of high quality. This raises a natural question: if pruning degrades a model’s performance, can retraining recover it? To explore this, we designed an experiment that contrasts a retraining scheme with a direct reference benchmark. %They are outlined in Table~\ref{tab:retraining_notation}.

The retraining scheme consists of progressively pruning a network during and after its training cycle. Starting from an initial dilution $p_0$, the model is trained normally and then subjected to an additional pruning step to reach a higher fraction $p>p_0$. This produces a deliberately degraded network—one whose performance has clearly deteriorated. We then retrain this pruned model to test whether it can recover its functionality or even surpass its previous performance despite starting from a compromised state. We call these {\it Retrained Models} and denote them as $(p_0,p)$, indicating the initial and additional pruning conditions.

As a benchmark for comparison, we introduce the {\it Reference Model}. Here, a new network is trained from scratch with a dilution $p_0=p$ imposed from the very beginning. This represents the straightforward route to obtaining a sparse model: training a network to adapt directly to the target level of sparsity throughout its entire learning process. We denote these reference models as $(p_0=p,0)$, meaning that the initial dilution was $p$ and that no further pruning was applied afterwards.

By comparing these two approaches, we aim to address a practical and conceptually important question: is it worth pursuing a strategy of progressive pruning followed by retraining, or is it more effective to simply train a new model with the desired sparsity from the outset? Understanding this trade-off is key to determining whether retraining offers any practical advantage or merely introduces unnecessary computational overhead with limited benefit. The outcome of this comparison sheds light on the influence of initial conditions on the learning dynamics of sparse networks and on the limits of retraining as a recovery strategy.

\begin{table}[htb!]
  \caption{Notation for the different network classes used in the retraining experiment.}
  \centering
  \begin{tabular}{lp{8cm}l}
    \toprule
    Notation & Description & Appearance in figures \\
    \midrule
    $(p_0, p)$ & \textbf{Retrained models}: networks first trained with pruning level $p_0$, then pruned further to $p$ and retrained. & Thick red \\
    $(p_0=p, 0)$   & \textbf{Reference models}: networks trained with a dilution $p$, without prior pruning or retraining. & Black \\
    — & \textbf{Frozen models}: networks trained with dilution $p_0$, pruned to $p$, but not retrained. & Yellow dots (in PCA) \\
    \bottomrule
  \end{tabular}
  \label{tab:retraining_notation}
\end{table}

\begin{figure*}[ht!]%
\centering
\begin{subfigure}{.45\textwidth}
  \includegraphics[keepaspectratio=true,width=0.9\textwidth]{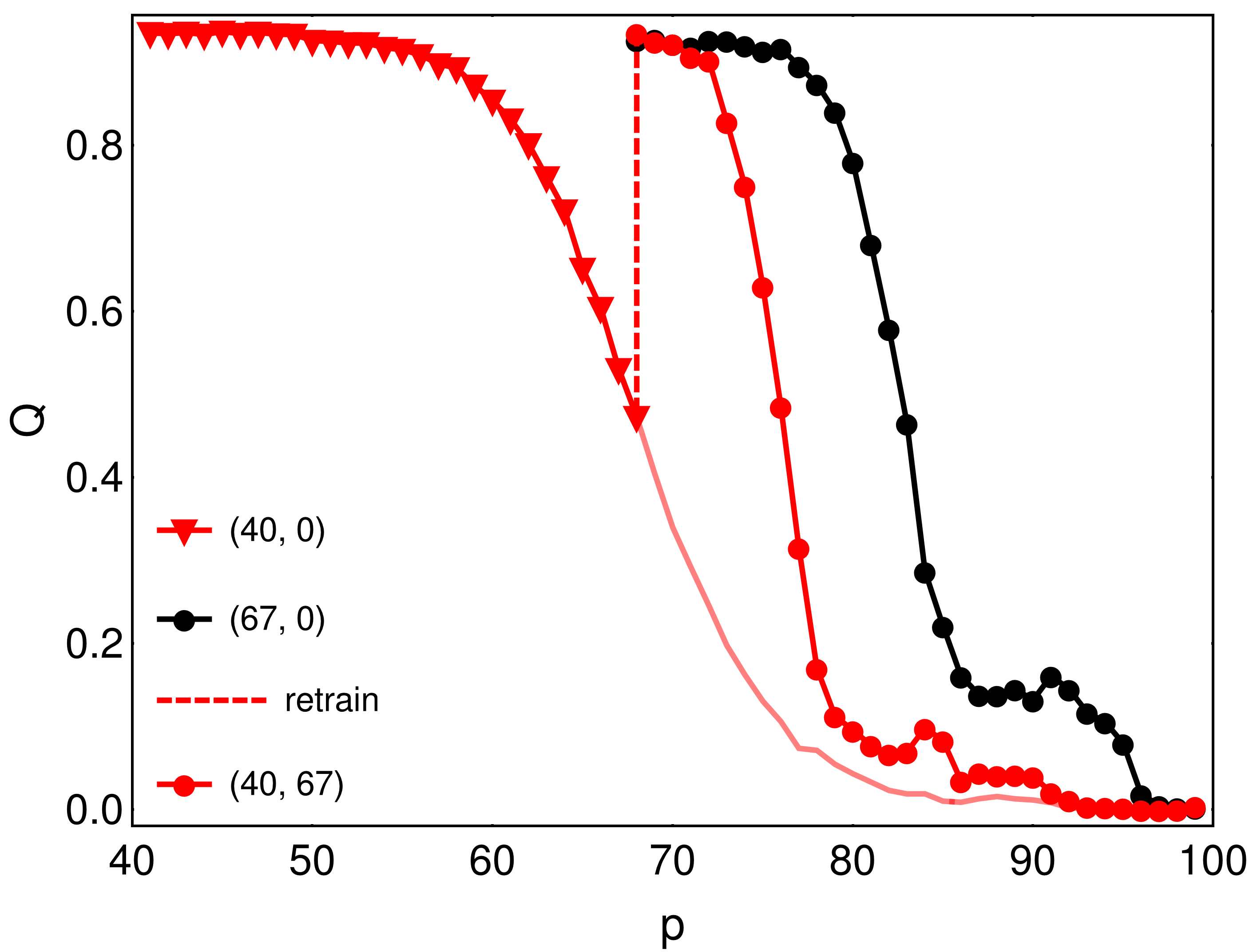}
  \caption{}
  \label{}
\end{subfigure}%
\begin{subfigure}{.45\textwidth}
  \centering
  \includegraphics[keepaspectratio=true,width=0.9\textwidth]{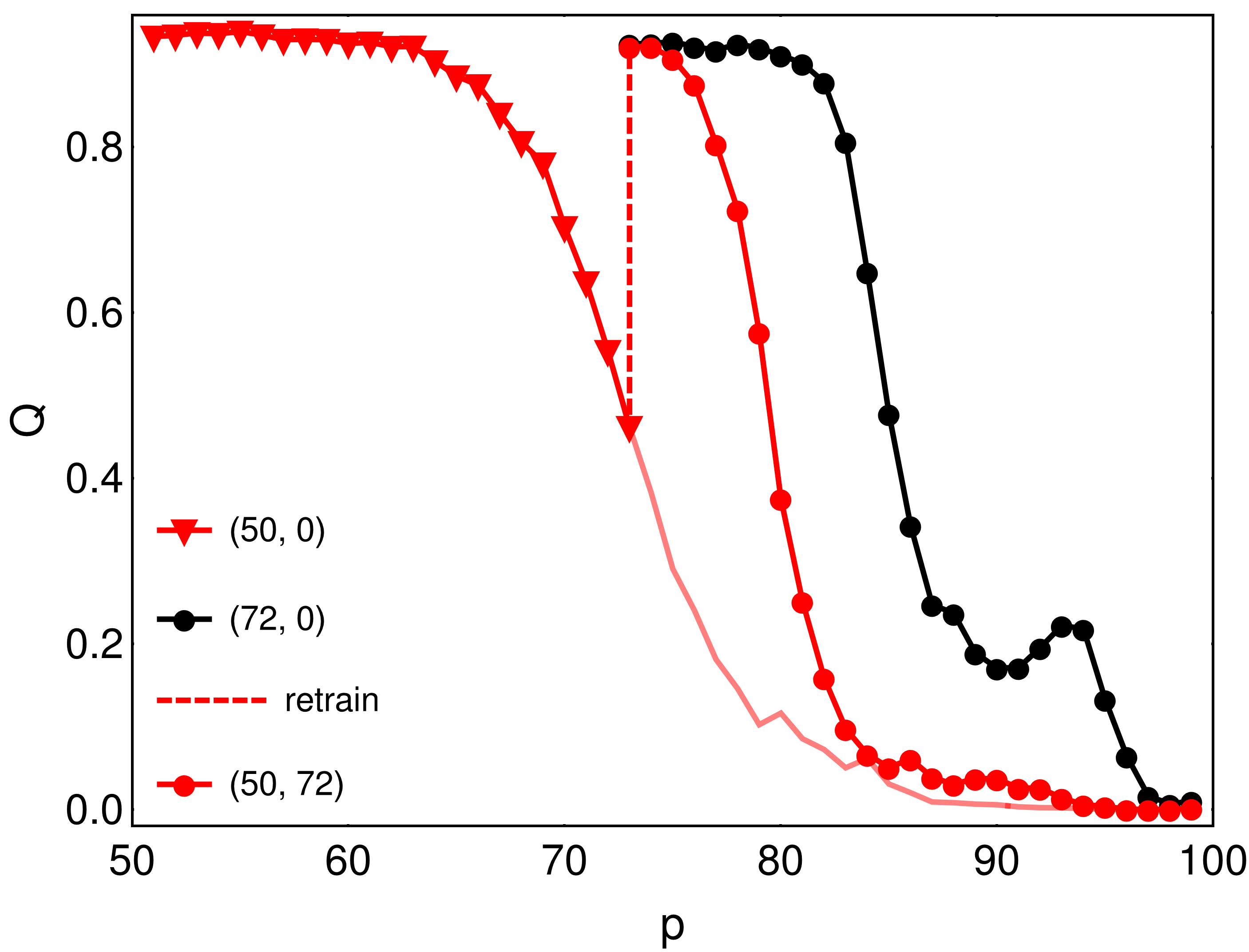}
  \caption{}
  \label{}
\end{subfigure}
\begin{subfigure}{.45\textwidth}
  \centering
  \includegraphics[keepaspectratio=true,width=0.9\textwidth]{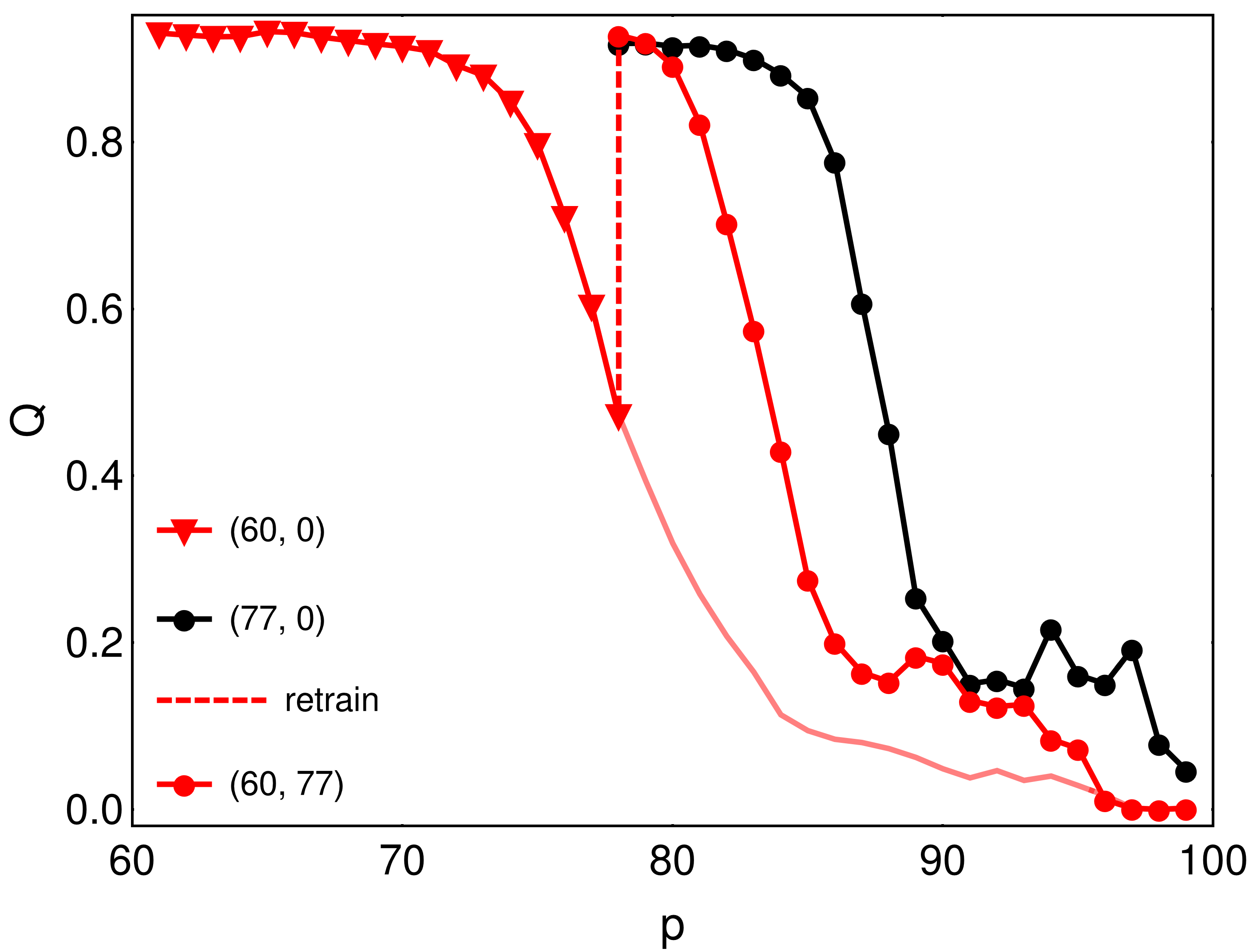}
  \caption{}
  \label{}
\end{subfigure}
\begin{subfigure}{.45\textwidth}
  \centering
  \includegraphics[keepaspectratio=true,width=0.9\textwidth]{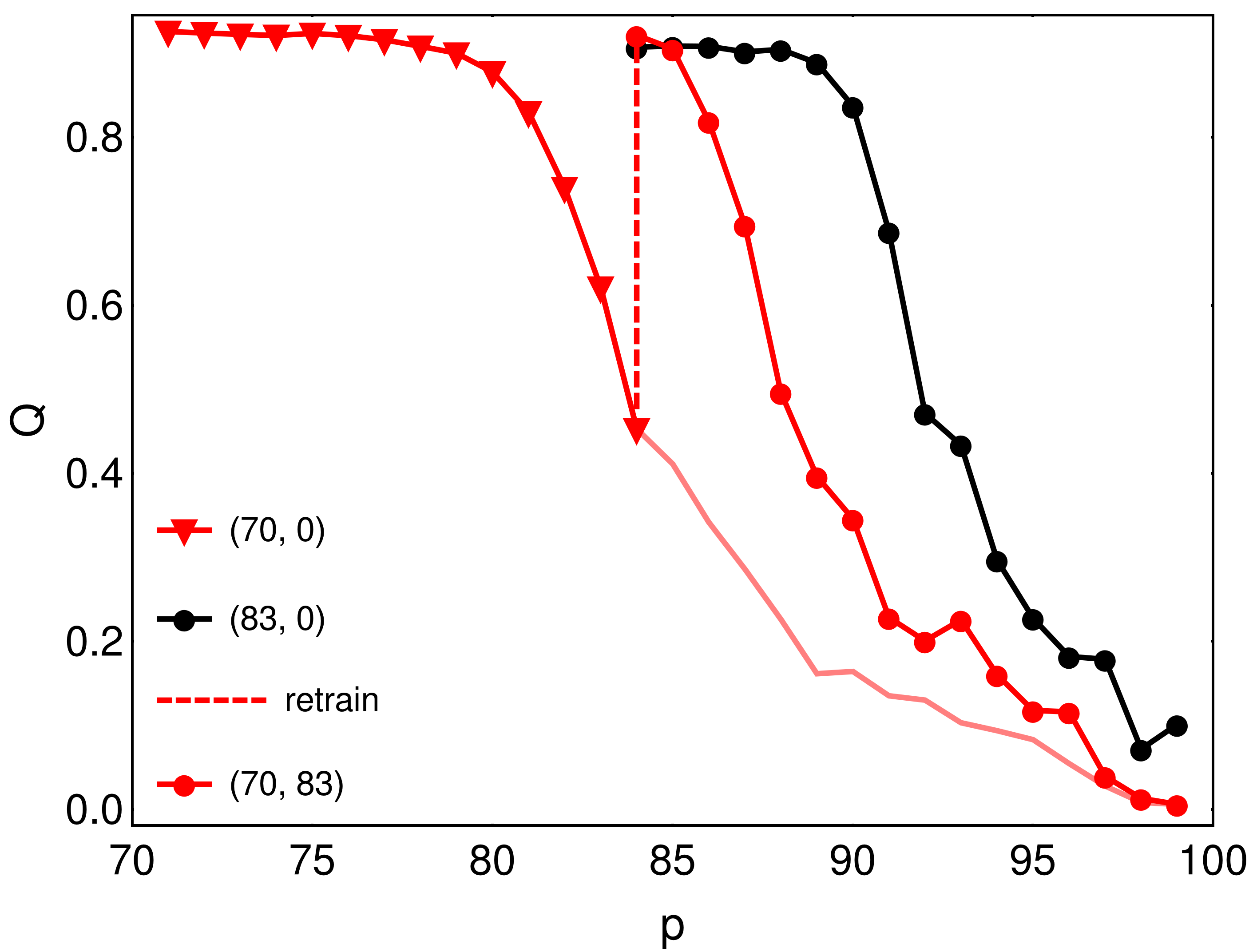}
  \caption{}
  \label{}
\end{subfigure}
  \caption{The quality of generation is graphed against the pruning of a network trained under initial conditions \( p_0 \) and retrained under \( p \) \( (p_0, p) \), and a network trained under initial conditions \( p \) \( (p, 0) \). This graph is shown for four different cases of \( p_0, p \). This allows for a comparison between the two training schemes. The generation quality of the \( (p_0, p) \) network always drops before that of the \( (p, 0) \) network.} 
  \label{retrain}
\end{figure*}

Figure~\ref{retrain} presents the results of both training strategies, with the reference models shown in black and the retrained models in red. The generation quality is plotted as a function of total pruning for both schemes. Four representative examples are shown for comparison. In all scenarios, the retrained networks initially recover high generation quality (close to 1). However, their performance deteriorates more rapidly as additional pruning is applied, consistently underperforming relative to the reference models. Although the difference is not large, it is systematic: in every case, the quality of the retrained models declines sooner and more sharply. This indicates that the additional pruning used to “break” the network introduces degradation that retraining cannot fully reverse.

\subsection{Comparing the models through generalized Ising models}
\begin{figure}[ht!]%
    \centering
    \begin{subfigure}[t]{0.45\linewidth}
        \centering
        \includegraphics[width=\linewidth]{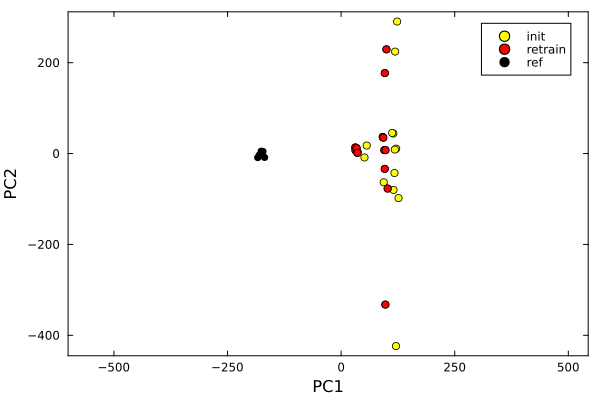}
          \caption{}
        \label{fig:pcaa}
    \end{subfigure}%
    \hfill
    \begin{subfigure}[t]{0.45\linewidth}
        \centering
        \includegraphics[width=\linewidth]{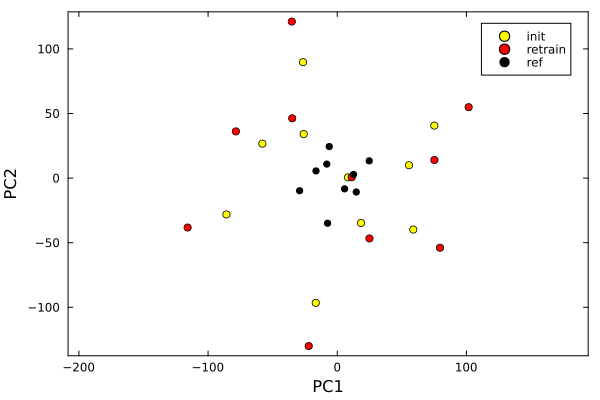}
        \caption{}
        \label{fig:pcab}
    \end{subfigure}

    \vspace{0.5cm}

    \begin{subfigure}[t]{0.45\linewidth}
        \centering
        \includegraphics[width=\linewidth]{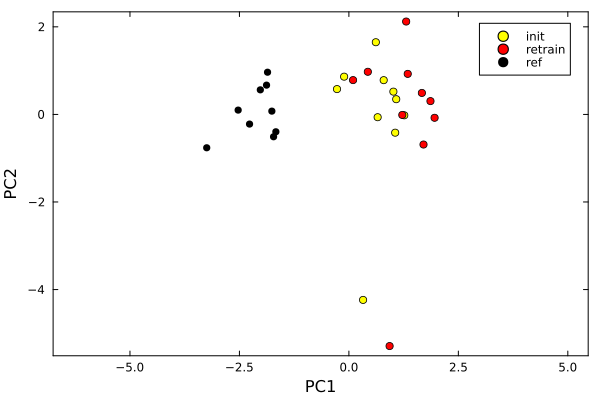}
        \caption{}
        \label{fig:pcac}
    \end{subfigure}%
    \hfill
    \begin{subfigure}[t]{0.45\linewidth}
        \centering
        \includegraphics[width=\linewidth]{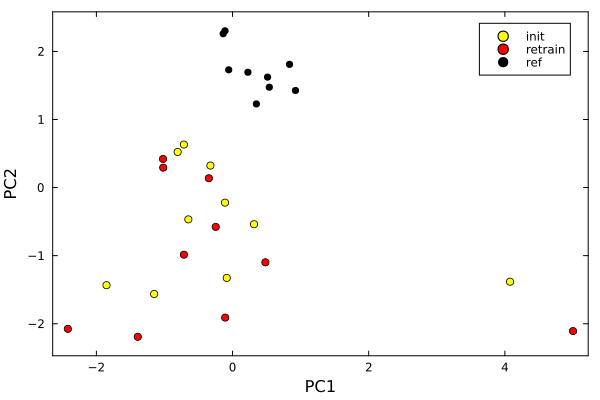}
        \caption{}
        \label{fig:pcad}
    \end{subfigure}

   \caption{Principal Component Analysis (PCA) of different network classes: reference models (black), retrained models (red), and frozen models (yellow). 
(a) PCA of the weight matrices, showing an apparent separation between reference models and the group of frozen and retrained models. 
(b) PCA of reference models trained with the same pruning mask as frozen and retrained ones, confirming that the separation in (a) is mainly due to the mask of zeros. 
(c) PCA of the effective two-body couplings obtained through the generalized Ising model (GIM) mapping, where frozen and retrained models still cluster together, distinct from reference models. 
(d) PCA of the couplings for reference models trained with the same pruning mask, showing that the clustering persists, indicating that retrained networks remain confined to regions of parameter space influenced by their initial training history.}
    \label{fig:4panels}
\end{figure}

To understand the origin of the differences between retrained and reference models, we hypothesized that retrained networks remain trapped in the ``broken'' conditions imposed at the moment of retraining. In contrast, reference networks, even with the same overall dilution, begin with fully random initializations for all active weights. To test this idea, we needed a framework where retrained, frozen, and reference models could be compared in a meaningful way based on their parameters.

A natural first step is to compare the weight matrices directly. One way to do this is to perform PCA on the weight matrices of multiple replicas from each class. Figure~\ref{fig:pcaa} shows such an analysis: two clusters emerge, one corresponding to the reference models and another including both frozen and retrained models. At first sight this seems to confirm our hypothesis, but the conclusion would be misleading. The observed separation is largely explained by the experimental design: frozen and retrained models share the same mask of zero weights by construction, whereas reference models do not. To verify this, we trained an additional set of reference models while enforcing the same pruning mask as in the frozen and retrained ones, but with randomly initialized nonzero weights. Figure~\ref{fig:pcab} shows the result: the clustering vanishes, confirming that the previous separation was driven by the pruning mask rather than by intrinsic similarities in the trained weights.

This highlights a deeper issue: direct comparison of weight matrices is problematic due to symmetries and redundancies in the RBM parameterization. In particular, many RBMs that differ at the microscopic parameter level can represent the same effective solution because of hidden-unit permutations and other equivalences. For this reason we consider PCA on raw weight matrices only as a naive first step. A more principled approach is to map each RBM to its corresponding generalized Ising model (GIM). Unlike the RBM weights, the GIM representation is unique: different RBM configurations, equivalent in terms of their distributions, map to the same GIM \cite{bea_infer_coup}. Given the weights $w$ and biases $b$, $c$, one can write the equivalence between the RBM Hamiltonian~\eqref{eq:hamiltonian} and the GIM Hamiltonian~\eqref{gim}

\begin{equation}
\mathcal{H}_{\mathrm{GIM}}(\bm{\sigma}) = - \sum_{j} H_j \sigma_j - \sum_{j_1>j_2} J^{(2)}_{j_1 j_2} \sigma_{j_1} \sigma_{j_2} 
- \sum_{j_1>j_2>j_3} J^{(3)}_{j_1 j_2 j_3} \sigma_{j_1} \sigma_{j_2} \sigma_{j_3} + \cdots,
\label{gim}
\end{equation}

where the $H_j$ are effective fields and the $J^{(k)}$ represent $k$-body couplings among spins. Previous works ~\cite{cossu2019,bulso2021, barra2012, feng2023} have derived explicit expressions for the effective couplings of the generalized Ising model in terms of the underlying RBM parameters. In particular in ~\cite{bea_infer_coup} , Decelle et al. obtained the following relations for the effective fields and two-body interactions:  

\begin{equation}
H_j = \eta_j + \frac{1}{2} \sum_i \mathbb{E}_{X_i^{(j)}} 
\left[
    \ln \frac{\cosh\left(\zeta_i + w_{ij} + X_i^{(j)}\right)}
             {\cosh\left(\zeta_i - w_{ij} + X_i^{(j)}\right)}
\right],
\label{1body}
\end{equation}

\begin{equation}
J_{j_1 j_2}^{(2)} = \frac{1}{4} \sum_i \mathbb{E}_{X_i^{(j_1 j_2)}} 
\left[
    \ln \frac{
        \cosh\!\left(\zeta_i + w_{ij_1} + w_{ij_2} + X_i^{(j_1 j_2)}\right)
        \cosh\!\left(\zeta_i - (w_{ij_1} - w_{ij_2}) + X_i^{(j_1 j_2)}\right)
    }{
        \cosh\!\left(\zeta_i + (w_{ij_1} - w_{ij_2}) + X_i^{(j_1 j_2)}\right)
        \cosh\!\left(\zeta_i - (w_{ij_1} + w_{ij_2}) + X_i^{(j_1 j_2)}\right)
    }
\right].
\label{2body}
\end{equation}

These expressions are obtained in the lattice-gas convention, where the spin variables take values $S_j \in \{0,1\}$ instead of the usual Ising convention $\sigma_j \in \{-1,+1\}$. In this framework, the parameters entering Eqs.~(\ref{1body})--(\ref{2body}) are defined as

\begin{equation}
\eta_j = \frac{1}{2}\!\left(b_j + \frac{1}{2}\!\sum_i W_{ij}\right), \qquad
\zeta_i = \frac{1}{2}\!\left(c_i + \frac{1}{2}\!\sum_j W_{ij}\right), \qquad
w_{ij} = \frac{1}{4}W_{ij},
\end{equation}
and the auxiliary variables $X_i^{(j)}$ and $X_i^{(j_1 j_2)}$ are defined as
\begin{equation}
X_i^{(j)} = \sum_{k \neq j} w_{ik}\sigma_k, \qquad
X_i^{(j_1 j_2)} = \sum_{k \neq j_1,j_2} w_{ik}\sigma_k,
\end{equation}
which represent the summed contributions from all other visible spins connected to hidden unit $i$. Under the central limit approximation, these variables are treated as Gaussian with zero mean and variance $\sum_k w_{ik}^2$.

In ~\cite{bea_infer_coup} the authors showed that this mapping captures the effective interactions of RBMs better than other approaches proposed in the literature. Because of this, we employed their formalism to infer the couplings of our models and to compare them using the same PCA methodology as before.

Figure~\ref{fig:pcac} shows the PCA projection of the inferred 2-body couplings for frozen, retrained, and reference models. A clustering pattern emerges that resembles the weight-based analysis in Figure~\ref{fig:pcaa}. To ensure that this effect was not simply due to mask structure, we repeated the analysis with reference models trained under the same pruning masks as the frozen and retrained ones (Figure~\ref{fig:pcad}). In this case, the clustering does not vanish: although the separation is less pronounced than in the weight-space analysis, retrained and frozen models still group together, distinct from the reference models. This indicates that retraining does not erase the imprint of early training conditions. Instead, networks remain confined to a region of parameter space determined by their initial training history. The observed performance drop is therefore not caused by the pruning mask alone, but by the specific weight values inherited from the first stage of training, values that retraining cannot fully escape.  

The PCA projection also reveals an interesting structure, particularly in how frozen and retrained models tend to occupy the same region. Since different replicas generally have distinct pruning masks, one might expect that clustering would occur only within individual pairs that share the same mask, rather than across the whole set. Yet, this is not what the analysis shows: frozen and retrained models cluster together more broadly, even beyond those specific pairs. This raises the question of what feature they all share that sets them apart from the reference models, even when the latter are trained with the same mask. The answer lies in how the additional pruning is applied. For frozen and retrained models, pruning always removes the weights with the smallest magnitudes after training, whereas in reference models the mask is imposed independently of the learned weight values. The observed separation in PCA space therefore reflects the similarities introduced by this targeted pruning, which systematically selects the weakest connections.

\section{Conclusions}

This work explored the resilience of Restricted Boltzmann Machines (RBMs) under pruning, giving insights into how neural networks adapt to sparsity. We found that RBMs can maintain strong generative performance even when a large majority of connections are pruned at initialization, demonstrating their ability to concentrate essential information into a small subset of weights. This aligns with broader observations in deep learning, where overparameterized models often rely on only a fraction of their parameters, and supports the Lottery Ticket Hypothesis. The sharp transition to failure under additional pruning demostrates the existence of a critical threshold, beyond which the network loses its ability to capture the data distribution. Notably, the relationship between initial and post-training pruning follows a simple scaling law, implying an underlying universality in how these models distribute learned information across their remaining connections.

%This work explored the resilience of Restricted Boltzmann Machines (RBMs) under pruning, giving insights into how neural networks adapt to sparsity. We found that RBMs can maintain strong generative performance even when a large majority of connections are pruned at initialization, demonstrating their ability to concentrate essential information into a small subset of weights. This aligns with broader observations in deep learning, where overparameterized models often rely on only a fraction of their parameters, and supports the Lottery Ticket Hypothesis by showing the existence of viable subnetworks in these generative architectures. The sharp transition to failure under additional pruning suggests the existence of a critical threshold, beyond which the network loses its ability to capture the data distribution. Notably, the relationship between initial and post-training pruning follows a simple scaling law, implying an underlying universality in how these models distribute learned information across their remaining connections.
The results of the retraining experiment directly contrast with the intuitive belief that additional training would improve the model's performance. Despite retraining restoring some functionality, the networks remained constrained by their initial learning trajectory, performing worse than models trained from scratch under the same sparsity conditions. This limitation appears rooted in the weight configurations inherited from early training, rather than the pruning structure itself. This finding underscores that the performance of a network is highly sensitive to its initial conditions, and once poorly learned, these weights are difficult to correct. The conclusion of this experiment is clear: to obtain a functional network with maximal sparsity, it is more effective to impose stringent initial dilution conditions, \( p_0 \), from the outset, rather than relying on a strategy of training, diluting, and retraining.

%The results of the retraining experiment directly contrast with the intuitive belief that additional training would improve the model's performance. Despite retraining restoring some functionality, the networks remained constrained by their initial learning trajectory, performing worse than models trained from scratch under the same sparsity conditions. This limitation appears rooted in the weight configurations inherited from early training, rather than the pruning structure itself, as resetting the weights before retraining fully restored performance. This finding underscores that the performance of a network is highly sensitive to its initial conditions, and once poorly learned, these weights are difficult to correct. The conclusion of this experiment is clear: to obtain a functional network with maximal sparsity, it is more effective to impose stringent initial dilution conditions, \( p_0 \), from the outset, rather than relying on a strategy of training, diluting, and retraining. In simple terms, what is poorly learned is hard to fix.

We expect that these results may extend beyond RBMs, and offers lessons for the design of efficient, sparse neural networks. The existence of a minimal "core" of weights necessary for generation suggests that careful initialization and progressive pruning may be more effective than aggressive post-hoc sparsification and retrainning. Moreover, the limitations of retraining underscore how early training dynamics create path dependencies that determine a network's ultimate capabilities. Future research would explore whether these observations hold for more complex architectures or if alternative training strategies might overcome these constraints. %Ultimately, this study advances our understanding of how neural networks balance efficiency and flexibility, while revealing the persistent influence of initial conditions on their learning potential. {\bf SHOULDN't WE SAY SOMETHING MORE GENERAL ABOUT PRETRAINNED MODELS?} {\color{red}I don't know, what do you suggest?} {\bf I'M NOT SURE, BUT PEOPLE IS USING A LOT OF PRETAINED MODELS NOW. WHAT DO THIS MEAN IN THIS CONTEXT?}

\section{Acknowledgments}
We will like to thank B. Seoane for valuable suggestions during the completion of this work. This work was supported by the Marie Sklodowska Curie Action (MSCA) Staff Exchange project “SIMBAD” (REA Grant Agreement n. 101131463)

\appendix
\section{Measuring Generation Quality}
\label{appendix:negative samples}

A key element of this work is the need to assess, in a systematic and reproducible way, the quality of the samples generated by the RBMs. As discussed throughout the main text, a successful generative model is one that produces samples with a distribution as close as possible to that of the original training data. In the case of MNIST, this means generating images that are diverse, clear, and recognizable as handwritten digits. However, relying on visual inspection to evaluate generation quality is neither objective nor scalable, especially when dealing with large numbers of replicas and generated samples.

To address this, we constructed an auxiliary evaluation tool with two main motivations: to emulate how a human observer might intuitively assess the quality of the generated digits, and to provide a computationally feasible method suitable for large-scale comparisons. This tool consists of an additional classifier specifically trained to distinguish between digit-like images and non-digit images. The classifier was trained in a supervised manner using a dataset composed of the MNIST digits as positive examples and a curated collection of $28\times28$ pixel images as negative examples. Constructing this negative dataset required particular care. While the concept of “digit” is well defined, the set of all possible “non-digit” images is practically infinite and highly heterogeneous. Moreover, if this set were sampled uniformly, the overwhelming majority of possible images would be completely random noise, potentially making the classifier trivial or biased.

\begin{figure}[ht!]%
    \centering
    \includegraphics[width=0.5\linewidth]{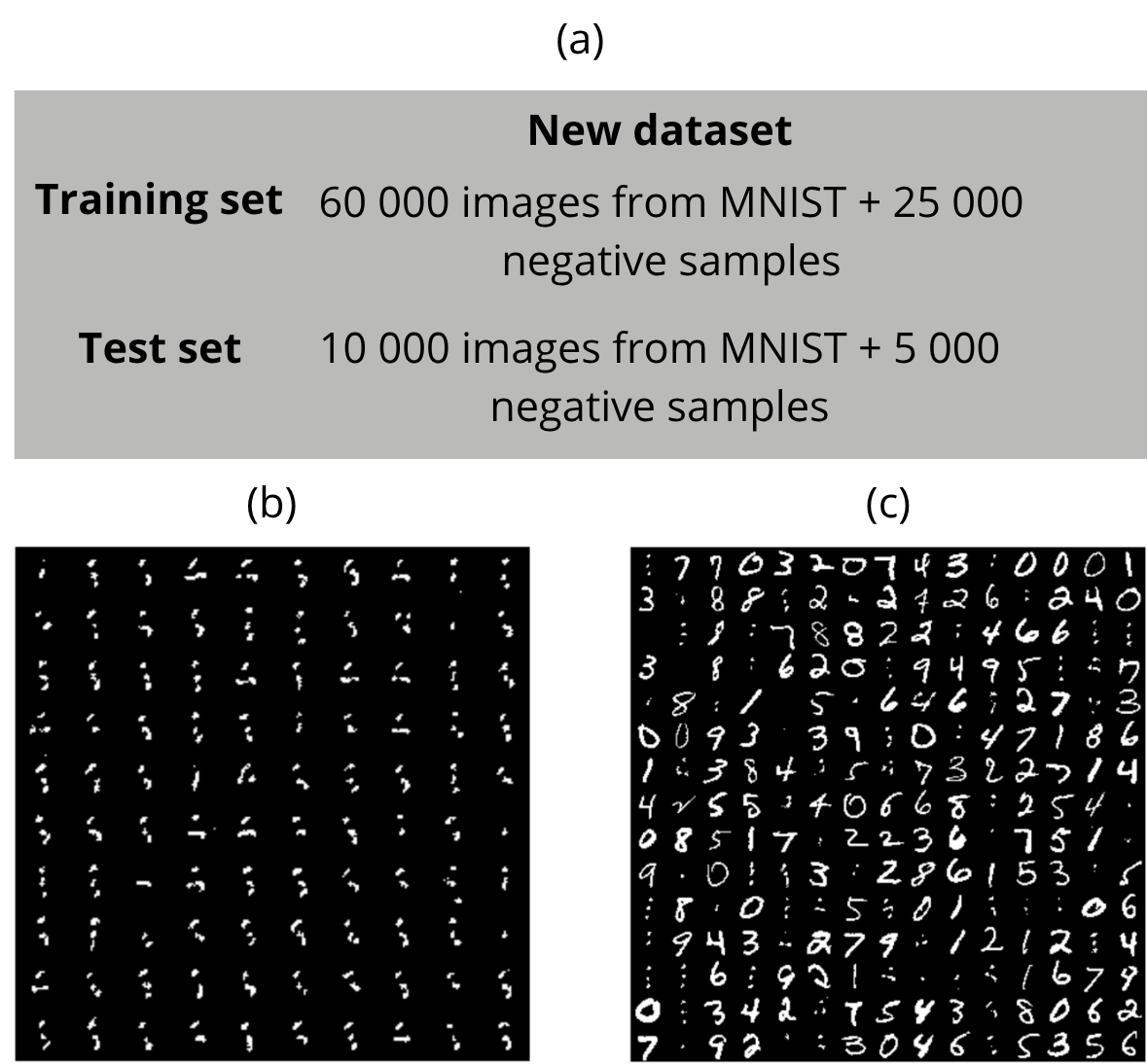}
    \caption{(a) Schematic representation of the training and test sets used to train the auxiliary classifier. The classifier was designed to first determine whether an image corresponds to a digit or not, and—if sufficiently confident—to further classify it into one of the 10 possible digit classes. (b) Examples of negative samples included in the dataset. These were extracted from a variety of RBMs trained and pruned to different levels, ensuring that the diverse “failure signatures” produced by distinct machines were well represented. (c) Sample of the final dataset used to train the auxiliary classifier, showing a mix of MNIST digits and negative examples.}
    \label{fig:negative samples}
\end{figure}

Fortunately, the specific context of our experiments simplifies the problem. As pruning levels increase, the typical failure mode of RBMs trained on MNIST is not the production of random noise but rather images that resemble sparse strokes over a black background, often centered on the canvas. This observation guided the construction of the negative dataset, which consists of variations on this kind of structured noise. The resulting classifier achieved an accuracy of 99\% on a dedicated test set.

The output of this model is a scalar $Q$ between 0 and 1, representing the classifier’s confidence that a given image corresponds to a digit. This value is used throughout the experiments as a proxy for generation quality. Nevertheless, it is important to recognize the limitations of this approach: when presented with examples that are very different from the training data, the classifier’s output may become unreliable. These so-called out-of-distribution or negative samples are an unavoidable challenge in this kind of evaluation.

In addition to $Q$, we introduced a second classifier, trained to recognize which specific digit is present in an image, provided that the first classifier has assigned it a sufficiently high confidence score. This allows us not only to measure the quality of individual samples, but also to evaluate the diversity of the generated dataset by tracking the frequency of each digit class. Both quantities—the average $Q$ and the frequency distribution of digits—are used to produce the generation quality and diversity plots (\ref{quality}, \ref{diversidad}, \ref{retrain}).

Although this evaluation scheme introduces an additional layer of black-box modeling and is inherently more heuristic, its results have consistently shown clear agreement with the other, more general and theoretically motivated metrics employed throughout this work. An additional advantage of this method is that, once the auxiliary classifier is trained, applying it to evaluate generation quality is computationally inexpensive compared to recomputing statistical divergences or adversarial distances. This makes it particularly practical when working with large numbers of generated samples or performing extensive parameter sweeps. While we remain cautious about overinterpreting the outputs of auxiliary models—especially outside the specific setup explored here—the observed agreement with the established metrics strengthens confidence in the overall evaluation framework and in the robustness of the conclusions drawn regarding the effects of pruning on learning and generation.

\end{document}